\documentclass[10pt, a4paper]{article}
\usepackage{lrec}
\usepackage{multibib}
\newcites{languageresource}{Language Resources}
\usepackage{graphicx}
\usepackage{tabularx}
\usepackage{soul}

\usepackage{epstopdf}
\usepackage[utf8]{inputenc}

\usepackage{hyperref}
\usepackage{xstring}

\usepackage{color}

\title{Dirichlet-Smoothed Word Embeddings for Low-Resource Settings}

\name{Jakob Jungmaier, Nora Kassner, Benjamin Roth}

\address{Center for Information and Language Processing, Ludwig-Maximilians-Universität München\\
         jakob.jungmaier@campus.lmu.de,
         \{kassner, beroth\}@cis.uni-muenchen.de\\}

\abstract{
	Nowadays, classical count-based word embeddings using positive pointwise mutual information (PPMI) weighted co-occurrence matrices have been widely superseded by machine-learning-based methods like word2vec and GloVe.
	But these methods are usually applied using very large amounts of text data. In many cases, however, there is not much text data available, for example for specific domains or low-resource languages. This paper revisits PPMI by adding Dirichlet smoothing to correct its bias towards rare words. We evaluate on standard word similarity data sets and compare to word2vec and the recent state of the art for low-resource settings: Positive and Unlabeled (PU) Learning for word embeddings.
	The proposed method outperforms PU-Learning for low-resource settings and obtains competitive results for Maltese and Luxembourgish.
	 \\ \newline  }

\begin{document}

\maketitleabstract

\section{Introduction}
Word embeddings that capture the meaning of words as vectors are the basis of much of the recent progress in natural language processing.
Nowadays, the classical count-based way to obtain such word embeddings from a corpus by using positive pointwise mutual information (PPMI) weighted co-occurrence matrices has been widely superseded by machine-learning-based methods like word2vec \cite{Mikolov_et_al_2013a,Mikolov_et_al_2013b} and GloVe \cite{Pennington_et_al_2014_(GloVe)}.
These methods are usually applied using very large amounts of text data.
But in many settings there is not much text data available, for example for specific domains or low-resource languages.

Recent investigations by \newcite{Konovalov_Tumunbayarova_2018} and \newcite{Jiang_et_al_2018} give rise to the conjecture that variants of the classical method to compute word embeddings might be more suitable for smaller corpora than the machine-learning-based methods. Additionally, \newcite{Levy_et_al_2015} suggest that certain choices of system design and hyperparameters for computing word embeddings can even be more impactful than the choice of the underlying model architecture itself.

In the spirit of these results, we propose to use a variant of the classical method for low-resource languages which tries to address a well-known problem of the PMI measure: its bias towards rare words \cite{Turney_Pantel_2010,Levy_et_al_2015,Jurafsky_Martin_2018}.
Following an idea mentioned by \newcite{Turney_Pantel_2010} and \newcite{Jurafsky_Martin_2018}, we apply \emph{Dirichlet smoothing} to weaken PMI's bias towards rare words and compare it to the word2vec skip-gram with negative sampling (SGNS) and Positive and Unlabeled (PU) Learning for word embeddings by \newcite{Jiang_et_al_2018}.

Models are trained on the English enwik9 corpus and evaluated using standard word similarity data sets. We investigate the models' performance for varying corpus size. In a case study, to demonstrate their practicability for low-resource languages, the three methods are used to train word embeddings for Maltese and Luxembourgish using respective Wikipedia dumps.

We show that our method i) outperforms strong baseline methods trained on enwik9 with biggest advantage on smaller corpora ii) scores best on the RW (Rare Word) similarity data set \cite{Luong_et_al_2013_(RW)} and iii) obtains outperforming and competitive results in the low-resource setting for Maltese and Luxembourgish.

We make our code publicly available.\footnote{\url{https://github.com/jungmaier/dirichlet-smoothed-word-embeddings}}

\section{Method}
Our proposed method adds Dirichlet smoothing to PPMI embeddings. To compute the standard PPMI embeddings, first, the corpus is scanned in windows consisting of a middle word and its surrounding context words and a word co-occurrence matrix is computed. Words that are frequent in the corpus cause high co-occurrence counts while more infrequent words (that might even be more informative) only result in very low counts. To prevent the overestimation of frequent words, the raw co-occurrence counts are substituted by \emph{pointwise mutual information (PMI)} \cite{Church_Hanks_1990}.
The PMI of a middle word $ w $ and a context word $ c $ is calculated as follows:
\begin{equation}\label{eq:pmi}
\textrm{PMI}(w,c) = \log \left( \frac{P(w,c)}{P(w) P(c)} \right)
\end{equation}
where $ P(w,c) $ is the probability that $ w $ and $ c $ occur in the same context (window) and $ P(w) $ and $ P(c) $ are the probabilities of the independent occurrences of the single words $ w $ and $ c $. These probabilities are estimated from the co-occurrence matrix by maximum likelihood estimation.

The numerator in equation (\ref{eq:pmi}) is the probability of a co-occurrence of the words given their actual distribution in the corpus (and some fixed window size), while the denominator is the expected probability that the two words co-occur given they were distributed independently across the corpus.

PMI faces a problem if two words do not co-occur at all. Then, the fraction will be zero and consequently $ \textrm{PMI}(w,c) = log(0) = -\infty $. 
A rather pragmatic solution is to simply leave these entries out and let them stay implicit zeros.
But this results in an inconsistent matrix since it attributes higher correlation (namely zero) to word pairs that never appeared than it does to word pairs that merely appeared less often than expected (negative values). However, it is still a convenient way of weighting a co-occurrence matrix that has proven to work well in practice \cite{Levy_Goldberg_2014}.

\newcite{Bullinaria_Levy_2007} demonstrated that the performance of the resulting vectors for word similarity tasks is better if all negatively correlated values are excised from the matrix. The result is called \emph{positive} pointwise mutual information:
\begin{equation}\label{eq:ppmi}
\textrm{PPMI}(w,c) = \max \left( \textrm{PMI}(w,c), 0 \right).
\end{equation}
The reason why PPMI performs better than PMI with respect to word similarity might be that usually the fact that two words appear less often together than expected does not convey much information. After all, even for humans it is very hard to name words that are negatively correlated, while it is relatively easy to name words that are positively correlated \cite{Levy_Goldberg_2014}.

(P)PMI suffers from a bias towards rare words \cite{Turney_Pantel_2010,Levy_et_al_2015,Jurafsky_Martin_2018}. The original intention to use PMI is to reduce the influence of the absolute frequency of words in the corpus, which also means to lower the weight of a co-occurrence with a more frequent word in comparison to the co-occurrence with a less frequent word. But one effect of this weighting procedure is that co-occurrences with rare words result in very high PMI values, which in turn overestimates the influence of co-occurrences with rare words.

\newcite{Turney_Pantel_2010} mention the following case: suppose that two words $ w $ and $ c $ are strongly statistically dependent, i.e., $ P(w,c) \approx P(w) \approx P(c) $. For example, this could be the case for a collocation like ``San Francisco''. Then, the PMI value of $ w $ and $ c $ is $ \log (1/P(w)) $ and consequently this value increases when the probability of the word $ w $ decreases. Now, supposed that ``San Francisco'' only appears, say, once in the corpus, the PMI value will be very high. But that in turn will skew the relation to the PMI values of other co-occurrences. As \newcite{Levy_et_al_2015} point out, this even creates a situation in which the top ``distributional features'' of a word, i.e., its context words, are often extremely rare words. But these do not necessarily appear in the respective representations of words that are semantically similar to that word. Moreover, the chance for the co-occurrence with a very rare word to be rather accidental than systematic is certainly higher than for co-occurrences with more frequent words. This means that (P)PMI might give very high weights to words that are in fact not significantly connected.

We approach the problem of rare words for PMI building on an idea mentioned by \cite{Turney_Pantel_2010} and \cite{Jurafsky_Martin_2018}, which is to use \emph{Dirichlet smoothing}.
Usually, Dirichlet smoothing is used to obtain non-zero probabilities for unseen events by adding a small pseudo count $ \lambda $ in each likelihood estimation.
For the present purpose, the idea is to add a small pseudo count to every entry in the co-occurrence matrix.
Since the same $ \lambda $ will be added to every entry, the probability of frequent co-occurrences will be lowered a bit while that of rare co-occurrences will be raised. The same will happen to the probabilities of single words.

To maintain the sparsity of the matrix, a smoothed version of PMI is computed for the non-zero entries only. After all, it is not the aim here to get probabilities for co-occurrences that did not appear in the training corpus.

Following this idea, the smoothed PMI$_{\lambda} $ will be computed as follows:
\begin{equation}
\label{eq:add-lambda-pmi}
\textrm{PMI}_{\lambda}(w,c) = \log \left( \frac{P_{\lambda}(w,c)}{P_{\lambda}(w) P_{\lambda}(c)} \right)
\end{equation}
where
\begin{equation}
\label{eq:add-lambda-pmi-p-w-c}
P_{\lambda}(w,c) = \frac{f(w,c) + \lambda}{\sum_{(w',c') \in V_{w} \times V_{c}} f(w',c') + \lambda |V_{w} \times V_{c}|}
\end{equation}
where $ f(w,c) $ denotes the frequency of the co-occurrence of $ w $ and $ c $, $ V_{w} $ the vocabulary of all middle words, and $ V_{c} $ the vocabulary of all context words.
To preserve a probability distribution, the counts also have to be adjusted for the probability calculations of the single words $ w $ and $ c $:
\begin{equation}
\label{eq:add-lambda-pmi-p-w-alt}
P_{\lambda}(w) = \frac{\sum_{c \in V_{c}} f(w,c) + \lambda |V_{c}|}{\sum_{(w',c') \in V_{w} \times V_{c}} f(w',c') + \lambda |V_{w} \times V_{c}|}
\end{equation}
\begin{equation}
\label{eq:add-lambda-pmi-p-c-alt}
P_{\lambda}(c) = \frac{\sum_{w \in V_{w}} f(w,c) + \lambda |V_{w}|}{\sum_{(w',c') \in V_{w} \times V_{c}} f(w',c') + \lambda |V_{w} \times V_{c}|}.
\end{equation}
Here, the $ \lambda $'s in the nominators have to be multiplied by $ |V_{w}| $ (respectively $ |V_{c}| $) since adding $ \lambda $ to every co-occurrence count in the matrix will raise the count of the single words by $ \lambda |V_{w}| $ (respectively $ \lambda |V_{c}| $).
Using these formulas it is now possible to ``pretend'' to add $ \lambda $ to every count while in fact only computing PMI$ _{\lambda} $ for the existing counts.
For the actual weighting of the co-occurrence matrix, only the positive PMI$ _{\lambda} $-values will be used, which yields PPMI$ _{\lambda} $.

The PPMI$_{\lambda}$ matrix is still a very large but sparse matrix. To obtain dense word embeddings with only few dimensions, we follow the usual way of using \emph{truncated singular value decomposition} (SVD) for dimensionality reduction:
\begin{equation}\label{eq:tsvd-trunc}
\mathbf{X}' = \mathbf{U}_{k} \mathbf{\Sigma}_{k} \mathbf{V}^{\mathsf{T}}_{k}
\end{equation}
where the original matrix $ \mathbf{X} $ is decomposed into orthogonal unit-length column matrices $ \mathbf{U} $, $ \mathbf{V} $, and the diagonal matrix $ \mathbf{\Sigma} $ of ordered singular values of $ \mathbf{X} $ \cite{Deerwester_1990,Levy_Goldberg_2014}. For the reduced matrix $ \mathbf{X}' $ only the largest $ k $ singular values in $ \mathbf{\Sigma} $ and the corresponding columns of $ \mathbf{U} $ and $ \mathbf{V} $ are considered.
We follow \newcite{Levy_et_al_2015} and use $ \mathbf{U}_{k} $ as our final word embedding matrix and dismiss $\mathbf{\Sigma}_{k} $.

We refer to the full approach from corpus to word embeddings as SVD-PPMI$ _{\lambda} $ in this paper.

\section{Experimental Setup}

We compared the proposed method SVD-PPMI$_{\lambda} $ to two baseline methods: word2vec SGNS \cite{Mikolov_et_al_2013a,Mikolov_et_al_2013b} and PU-Learning for word embeddings \cite{Jiang_et_al_2018}.

\subsection{Corpora}
\label{subsec:corpora}
First, we train SVD-PPMI$ _{\lambda} $, word2vec SGNS, and PU-Learning on \emph{enwik9} which consists of the first $ 10^{9} $ bytes of an English Wikipedia dump from 2006, provided for download by Matt Mahoney.\footnote{\url{www.cs.fit.edu/~mmahoney/compression/textdata.html}} Removal of the markup language and further preprocessing was done by a Perl script by Mahoney to be found on the same web page.
Punctuation marks were removed completely, all letters were lowercased and words tokenized by whitespace. The resulting text file contains 124,301,827 tokens in total with a vocabulary of 833,185 words.

To see the influence of the corpus size on the quality of the trained word embeddings, we used differently sized subsets of the enwik9 corpus. For the following experiments, we used the first 1, 2, 4, 8, 16, 32, and 64 million words.

Experiments for Luxembourgish and Maltese were conducted using dumps of the Luxembourgish and Maltese Wikipedias from 2019. The complete dumps were preprocessed in the same manner as for the enwik9 corpus. The resulting corpus for Luxembourgish contains 6,268,907 tokens with a vocabulary size of 283,168. The resulting corpus for Maltese contains 1,617,402 tokens with a vocabulary size of 87,902.

\subsection{Evaluation}
\label{subsec:evaluation}

The final word embeddings are evaluated on five word similarity data sets:
\emph{RG-65} \cite{Rubenstein_Goodenough_1965_(RG-65)}, \emph{WordSim-353} \cite{Finkelstein_et_al_2001_(wordsim353)}, \emph{SimLex-999} \cite{Hill_et_al_2014_(SimLex-999)}, \emph{MEN} \cite{Bruni_et_al_2014_(MEN)}, and the \emph{RW (Rare Word)} data set by \newcite{Luong_et_al_2013_(RW)}.
Each of these data sets contains a number of word pairs with a corresponding gold standard similarity score assigned by human annotators. For every word pair in a data set, the cosine similarity of the corresponding word embeddings is computed. For the case that some word occurs in a word pair for which no corresponding word embedding was trained, the similarity for the pair will be set to zero. The final score is Spearman's rank correlation coefficient (Spearman's $ \rho $) of the scores assigned by humans and the cosine similarities of the corresponding word embeddings.

In a similar manner as \newcite{Jiang_et_al_2018} obtain data sets for languages other than English, data sets for Luxembourgish and Maltese were obtained for RG-65, WordSim-353, SimLex-999, and MEN via the Google translation API.\footnote{\url{https://cloud.google.com/translate}} Word pairs containing multi-word expressions after translation were removed. Occasionally, pairs with very similar words in the original English data sets resulted in translated pairs of twice the same word. Respective pairs were discarded as well.
Manual inspection of the final word pairs indicates that a small proportion of terms seem to remain English after translation, but this is not considered to be a problem since, after all, the conditions are equal for each approach tested here.\footnote{However, since this effect seemed to be particularly pronounced for the RW data set, no translations of this data set were used here.}
Table \ref{tab:datasets} shows the used data sets and their respective sizes.
\begin{table}[ht]
			\begin{tabular}{|l|ccc|}
			\hline
			Data set & English  & Luxembourgish & Maltese \\
			\hline
			RG-65 & 65 & 61 & 62 \\
			WordSim-353 & 353 & 342 & 338 \\
			SimLex-999 & 999 & 937 & 960 \\
			MEN & 3000 & 2904 & 2852 \\
			RW & 2034 & --- & --- \\
			\hline
			\end{tabular}
		\caption{Overview of the used word similarity data sets and their sizes (number of word pairs).}
		\label{tab:datasets}
\end{table}

\subsection{Implementation and Hyperparameter Choices}

For the comparison with the two baseline methods, the original implementations by the authors were used, i.e., for word2vec SGNS the original C-implementation by Mikolov et al.\footnote{\url{https://github.com/tmikolov/word2vec}} and for PU-Learning for word embeddings the original code provided by Jiang et al.\footnote{\url{https://github.com/uclanlp/PU-Learning-for-Word-Embedding}}

For all compared methods the window size was set to 5, and the minimum count for words was set to 1, i.e., representations were trained for all words in the corpora. %
The length of all word embeddings was set to the default word2vec dimension of 100.

For the  model-specific hyperparameters of the baseline methods the respective default values of the provided implementations were used. The $ \lambda $-parameter of our own model was selected based on the comparison of different values described in section \ref{subsec:influence-of-the-lambda-parameter}

\section{Results and Discussion}

Table \ref{tab:all-languages} summarizes Spearman's rank correlation coefficient for all similarity test sets. Trained on enwik9, our proposed method outperforms the two baseline methods on all similarity test sets.

\begin{table*}[ht]
	\begin{center}
		\begin{tabular}{|l|l|ccccr|}
			\hline
			Language & Method & RG-65 & WordSim-353 & SimLex-999 & MEN & RW \\
			\hline
			English & SGNS & .718 & .674 & .306 & .690 & .294 \\
			& PU-Learning & .700 & .689 & .267 & .718 & .217 \\
			 & SVD-PPMI$_{\lambda } $ & \textbf{.744} & \textbf{.738} & \textbf{.314} & \textbf{.726} & \textbf{.308} \\
			\hline
			Luxembourgish & SGNS & .146 & .133 & .069 & .186 & --- \\
			& PU-Learning & .267 & \textbf{.239} & \textbf{.075} & .337 & --- \\
			& SVD-PPMI$_{\lambda } $ & \textbf{.302} & .193 & .054 & \textbf{.351} & --- \\
			\hline
			Maltese & SGNS & .063 & .047 & .074 & .168 & ---\\
			& PU-Learning & .059 & .191 & .068 & .320 & ---\\
			& SVD-PPMI$_{\lambda } $ & \textbf{.141} & \textbf{.208} & \textbf{.106} & \textbf{.351} & --- \\
			\hline
		\end{tabular}
		\caption{Comparison of SGNS (word2vec), PU-Learning, and SVD-PPMI$_{\lambda }$ (ours) concerning word similarity tasks for English, Luxembourgish, and Maltese. Spearman's $ \rho $ between the human annotated gold standard and the cosine similarity scores are shown. Embeddings are trained on enwik9 and 2019 dumps of the Luxembourgish and Maltese Wikipedias respectively.}
		\label{tab:all-languages}
	\end{center}
\end{table*}

In the low-resource setting, SVD-PPMI$_{\lambda} $ outperforms baselines for Maltese and shows competitive performance for Luxembourgish embeddings.

\subsection{$ \lambda $-Parameter}
\label{subsec:influence-of-the-lambda-parameter}

\begin{figure*}[!h]
	\begin{center}
		\includegraphics[scale=0.4, trim=25 10 25 50,clip]{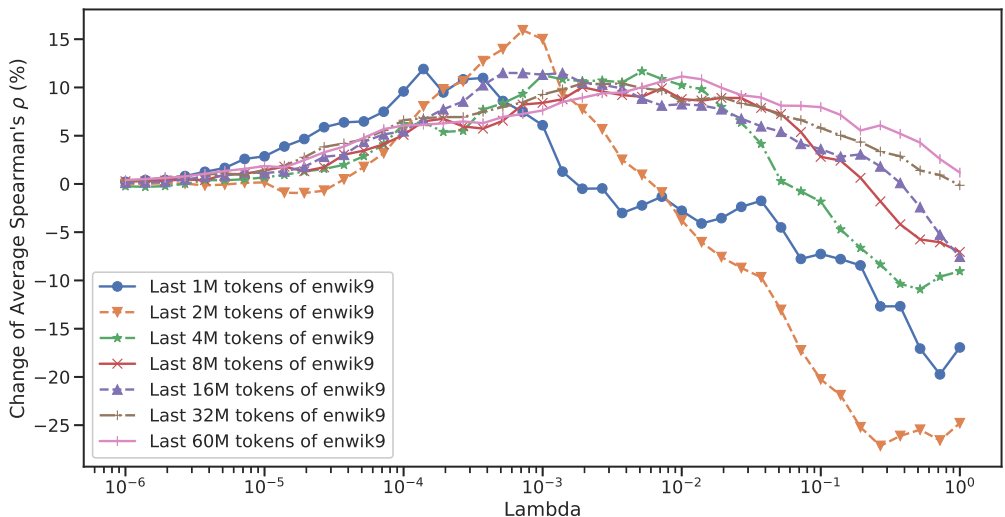} 
		\caption{Influence of the $ \lambda $-parameter on the average score (Spearman's $ \rho $) of all five used word similarity data sets (RG-65, WordSim-353, SimLex-999, MEN, RW) for different corpus sizes comparing 43 values of $ \lambda $ from $ 10^{-6} $ to $ 1 $.}
		\label{fig:lambda-comparison}
	\end{center}
\end{figure*}

The influence of the $ \lambda $-parameter on the performance of SVD-PPMI$ _{\lambda} $ is shown by a comparison of the average performance on all five word similarity data sets. $ \lambda $ is increased from $ 10^{-6} $ to $ 1 $. To see the influence of the corpus size and to select the best $ \lambda $'s for the following experiments, this comparison was made for the \emph{last} 1, 2, 4, 8, 16, 32, and 60 million tokens of enwik9. The results can be seen in figure \ref{fig:lambda-comparison}.
It shows that a good choice for $ \lambda $ can increase the performance by around 10 \% (for the last 8M tokens) to 16 \% (for the last 2M tokens). A $ \lambda $ chosen too high on the other hand can drop the performance considerably, as can be seen in the case of $ \lambda \approx 0.3 $ for the last 2M tokens, which decreases the score by around 27 \%. However, in most cases it seems that a $ \lambda $ chosen too low does not cause a notable drop in performance. This yields that, if in doubt, it might be advisable to stick to a smaller $ \lambda $. For increasing corpus size, the optimal $ \lambda $-value also has the tendency to increase.

For the experiments in the present paper, the best $ \lambda $-values of the last 1M tokens were used for training word embeddings for the \emph{first} 1M tokens of enwik9, the best $ \lambda $-values of the last 2M tokens were used for training word embeddings for the \emph{first} 2M tokens of enwik9 and so on. The best $ \lambda $ for the last 60 million tokens was used for word embeddings for both, the first 64 million tokens of enwik9 as well as the complete enwik9 corpus.

In the case of Maltese and Luxembourgish, the best $ \lambda $-values were chosen conservatively according to the best values for the next smaller tested corpora, i.e., 
the best $ \lambda $ for the last 1M tokens of enwik9 for Maltese and that for the last 4M tokens of enwik9 for Luxembourgish.

\subsection{Corpus Size}
\begin{figure*}[!h]
	\begin{center}
		\includegraphics[scale=0.6]{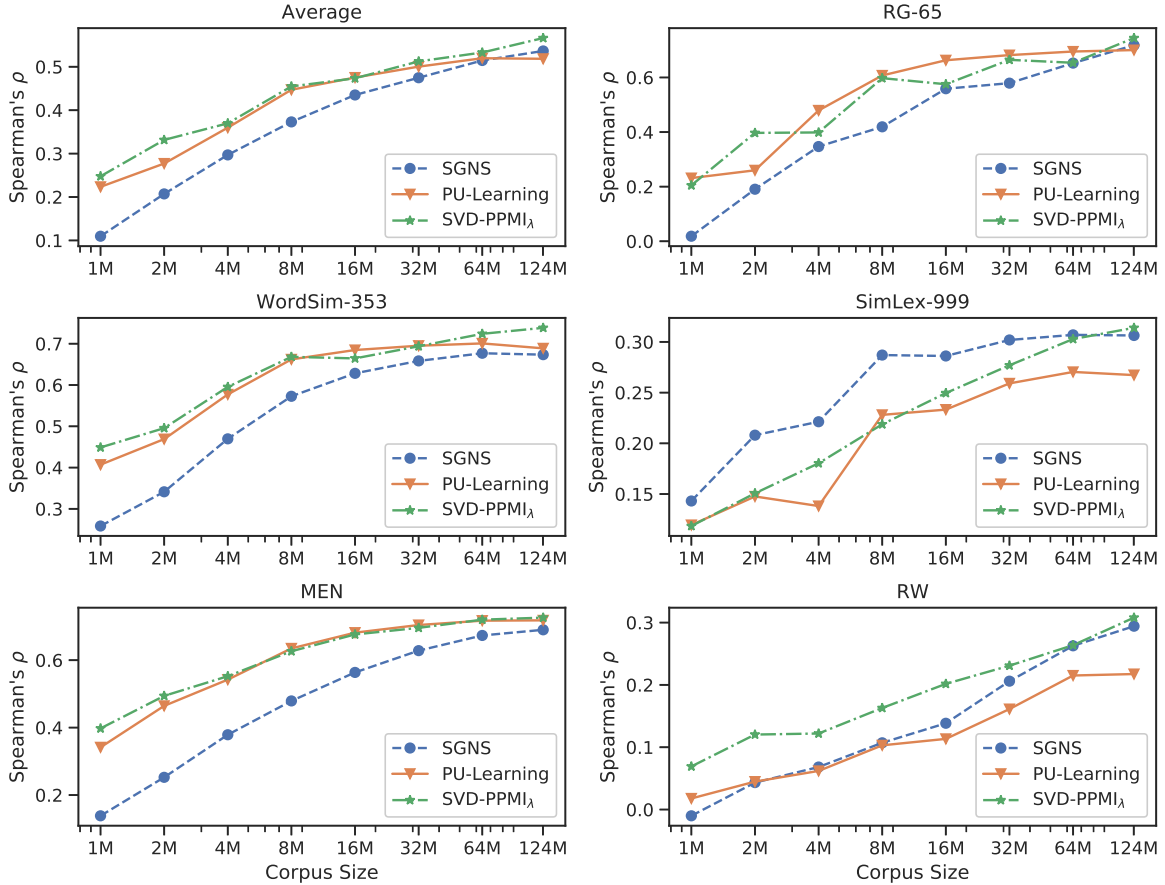} 
		\caption{Performance of the methods on word similarity tasks as a function of the corpus size (for differently sized slices of enwik9). The top left graph shows the macro average of all five used word similarity data sets.}
		\label{fig:wordsim-multi}
	\end{center}
\end{figure*}

The results of the comparison of SGNS, PU-Learning for word embeddings, and the present approach SVD-PPMI$_{\lambda}$ for different corpus sizes are shown in figure \ref{fig:wordsim-multi}.

It can be seen that SVD-PPMI$_{\lambda}$ outperforms the baseline approaches for most corpus sizes. Especially for the smaller sized corpora consisting of the first 1 and 2 million tokens of enwik9, SVD-PPMI$_{\lambda}$ yields better results. For bigger corpora it performs very similarly to the PU-Learning approach. Taking a look at the performance on the RW (Rare Word) data set reveals that SVD-PPMI$_{\lambda}$ seems to work particularly well for representing rare words.

\section{Related Work}

\subsection{Word Embeddings for Low-Resource Languages}

Only few explicit approaches have been made to learn word embeddings for low-resource languages from scratch, i.e., without trying to project existing embeddings from other languages into the source language.
\newcite{Konovalov_Tumunbayarova_2018} investigated the training of word embeddings for the mongolic language of Buryat using the classical way of factorizing a PPMI matrix by truncated SVD.
\newcite{Jiang_et_al_2018} proposed to learn word embeddings for low-resource languages from PPMI matrices by applying PU-Learning to overcome the sparsity of matrices caused by the lack of data. Neither of these methods used Dirichlet smoothing for PPMI.

\subsection{Smoothing the PMI Matrix}

\newcite{Levy_et_al_2015} approach PMI's bias towards rare words by a method called \emph{context distribution smoothing}, which is inspired by the choice of the noise distribution used by word2vec SGNS to generate negative samples.
The idea is to substitute PMI by
\begin{equation}\label{eq:cds}
\textrm{PMI}_{\alpha}(w,c) = \log \left( \frac{\hat{P}(w,c)}{\hat{P}(w)\hat{P}_{\alpha}(c)} \right)
\end{equation}
where $ \hat{P}_{\alpha}(c) $ is the smoothed context probability\
\begin{equation}\label{eq:cds-p-c}
\hat{P}_{\alpha}(c) = \frac{f(c)^{\alpha}}{\sum_{c' \in V_{c}} f(c')^{\alpha}}.
\end{equation}
The effect of this smoothing technique is that, given $ c $ is rare, the probability of the context word will be higher than before, i.e., $ \hat{P}_{\alpha}(c) > \hat{P}(c) $. This, in turn, reduces the PMI of co-occurrences with rare words.

Another possibility, proposed by \newcite{Pantel_Lin_2002}, is to multiply the PMI value by the following discount factor:
\begin{equation}
\label{eq:discount}
\delta_{w,c} = \frac{f(w,c)}{f(w,c) + 1} \cdot \frac{\min \left(  f(w), f(c) \right) }{\min \left( f(w), f(c) \right) + 1}.
\end{equation}
This causes that, the less frequent one of $ w $ or $ c $ gets, the more the final weight PMI$ _{\delta}(w,c) $ will be reduced. The left factor in the equation causes a similar reduction if the co-occurrence count of the word pair is low. All in all, $ \delta $ pushes the PMI values towards zero, more for rare words and less for frequent words or co-occurrences.

\newcite{Turney_Pantel_2010} and \newcite{Jurafsky_Martin_2018} mention the idea to use Dirichlet smoothing to weaken PMI's bias towards rare words.
\newcite{Turney_Littman_2003} used additive smoothing in the context of calculating association strength in order to avoid division by zero in their specific setting.
But, to the best of our knowledge, the idea of factorizing Dirichlet-smoothed count matrices for obtaining word embeddings
has not been carried out nor evaluated in previous work.

\section{Conclusion}
This work investigates classical PPMI embeddings with Dirichlet smoothing to correct its bias towards rare words. We show that classical PPMI based word embeddings can outperform machine-learning-based methods in a low-resource setting.

In a case study we demonstrated this on the low-resource languages Maltese and Luxembourgish. Further work should investigate its performance in domain-specific low-resource settings.

\section{Acknowledgements}
This work has been funded by the German Federal Ministry of Education and
Research (BMBF) under Grant No. 01IS18036A. The authors of this work take full
responsibilities for its content.

\section{Bibliographical References}
\label{main:ref}

\bibliographystyle{lrec}
\bibliography{CL_bib}

\end{document}